\documentclass[letterpaper]{article} 
\usepackage{aaai25}  
\usepackage{times}  
\usepackage{helvet}  
\usepackage{courier}  
\usepackage[hyphens]{url}  
\usepackage{graphicx} 
\usepackage{natbib}  
\usepackage{caption} 
\usepackage{algorithm}
\usepackage{algorithmic}

\usepackage{amsfonts}
\usepackage{amsmath}
\usepackage{array}
\usepackage{booktabs}
\usepackage{tabularx}

\usepackage{subfigure}
\usepackage{subcaption}

\usepackage{appendix}

\usepackage{amsthm}
\newtheorem{example}{Observation}

\usepackage{xcolor}

\usepackage{newfloat}
\usepackage{listings}
\DeclareCaptionStyle{ruled}{labelfont=normalfont,labelsep=colon,strut=off} 
\floatstyle{ruled}
\newfloat{listing}{tb}{lst}{}
\floatname{listing}{Listing}
\title{FedSA: A Unified Representation Learning via Semantic Anchors for Prototype-based Federated Learning}
\author{
    Yanbing Zhou\textsuperscript{\rm 1},
    Xiangmou Qu\textsuperscript{\rm 2},
    Chenlong You\textsuperscript{\rm 1},
    Jiyang Zhou\textsuperscript{\rm 1},
    Jingyue Tang\textsuperscript{\rm 1},
    Xin Zheng\textsuperscript{\rm 1},
    Chunmao Cai\textsuperscript{\rm 1,3},
    Yingbo Wu\textsuperscript{\rm 1}\thanks{Prof. Yingbo Wu is the corresponding author.}
}
\affiliations{
    \textsuperscript{\rm 1}Chongqing University\\
    \textsuperscript{\rm 2}OPPO Research Institute\\
    \textsuperscript{\rm 3}Chongqing Changan Automobile Co., Ltd\\
    \{202124021011, wyb\}@cqu.edu.cn, lokinko@oppo.com, \{ycl\}@stu.cqu.edu.cn
    
}

\usepackage{bibentry}

\begin{document}

\maketitle

\begin{abstract}
Prototype-based federated learning has emerged as a promising approach that shares lightweight prototypes to transfer knowledge among clients with data heterogeneity in a model-agnostic manner.
However, existing methods often collect prototypes directly from local models, which inevitably introduce inconsistencies into representation learning due to the biased data distributions and differing model architectures among clients.
In this paper, we identify that both statistical and model heterogeneity create a vicious cycle of representation inconsistency, classifier divergence, and skewed prototype alignment, which negatively impacts the performance of clients.
To break the vicious cycle, we propose a novel framework named \textbf{Fed}erated Learning via \textbf{S}emantic \textbf{A}nchors (FedSA) to decouple the generation of prototypes from local representation learning.
We introduce a novel perspective that uses simple yet effective semantic anchors serving as prototypes to guide local models in learning consistent representations.
By incorporating semantic anchors, we further propose anchor-based regularization with margin-enhanced contrastive learning and anchor-based classifier calibration to correct feature extractors and calibrate classifiers across clients, achieving intra-class compactness and inter-class separability of prototypes while ensuring consistent decision boundaries.
We then update the semantic anchors with these consistent and discriminative prototypes, which iteratively encourage clients to collaboratively learn a unified data representation with robust generalization.
Extensive experiments under both statistical and model heterogeneity settings show that FedSA significantly outperforms existing prototype-based FL methods on various classification tasks.
\end{abstract}

%

\section{Introduction}
Federated Learning (FL) has been proposed to collaboratively train a global model to address the increasingly significant privacy concerns \cite{li2020federated,kairouz2021advances}.
Despite its advantages, traditional FL methods \cite{mcmahan2017communication} struggle with \textit{statistical heterogeneity}, where heterogeneous data among clients can bias the global model towards clients with dominant data characteristics \cite{wang2020tackling}.
To remedy this, personalized FL \cite{tan2022towards} aims to train a personalized model for each client by leveraging the benefits of FL rather than pursuing a global model.
Many researchers have concentrated on model-based FL approaches that split the model into a body for universality and a head for personalization, requiring the same model architectures across all clients to facilitate the aggregation of the body at the server \cite{oh2021fedbabu}.
However, \textit{model heterogeneity} frequently arises when clients may design their own local model architectures to meet individual requirements and hardware constraints.
Such model-based FL approaches not only incur significant communication costs but also risk exposing model details, which may further raise concerns about privacy and commercial proprietary information \cite{ye2023heterogeneous,wang2023model}.

To address these issues, knowledge-based FL approaches have emerged as a novel FL paradigm that transfers various types of global knowledge among clients with heterogeneous data and diverse model architectures \cite{tan2022fedproto}.
For example, knowledge distillation (KD)-based FL methods transfer logits outputs from a teacher model as global knowledge to instruct the student models \cite{li2019fedmd, lin2020ensemble,zhang2021parameterized}.
However, these methods require a public dataset to align the outputs and are highly dependent on the quality of this dataset \cite{zhang2023towards}.
Researchers further explore KD in a data-free manner by employing additional auxiliary models as global knowledge \cite{wu2022communication,zhang2022fine}, but the communication costs for transmitting the auxiliary models remain considerable.
Recently, prototype learning has garnered increasing interest due to its exemplar-driven nature and intuitive interpretation \cite{snell2017prototypical}.
Prototypes, which are essentially averages of class representations, serve as abstract concepts that can effectively integrate feature representations from diverse data distributions.
By transferring lightweight prototypes as global knowledge, prototype-based FL methods significantly address privacy concerns and reduce communication costs \cite{tan2022fedproto,tan2022federated,zhang2024fedtgp}.

However, existing prototype-based FL methods directly collect prototypes from biased data distributions of clients, which inherently introduces inconsistencies into representation learning \cite{zhou2023fedfa}.
Moreover, the prototypes extracted from different model architectures have diverse scales and separation margins \cite{zhang2024fedtgp}, further exacerbating these inconsistencies.
These issues demonstrate that the traditional prototype generation is sub-optimal for heterogeneous FL since prototypes are heavily bound to the representation learning process and data distribution across clients, which can potentially lead to learning collapse \cite{ge2024beyond}.
To investigate the impact of representation inconsistencies on the generation of prototypes, we revisit and reinterpret the biases arising from heterogeneity in feature extractors and classifiers, from the perspective of representation learning \cite{guo2023fedbr}:
1) Biased feature extractors tend to produce inconsistent data representations in the semantic space;
2) Biased classifiers often diverge, learning skewed decision boundaries due to local class distributions;
3) Skewed prototype alignment occurs as local prototypes, generated from these inconsistent representations, naturally exhibit a skewed alignment in the semantic space.
The naive averaging aggregation of these skewed local prototypes \cite{tan2022fedproto} leads to the margin shrink problem, where the global prototypes tend to converge closely together, reducing the overall separability and effectiveness in distinguishing between different classes \cite{zhang2024fedtgp}.
Moreover, the global prototypes then serve to guide the representation learning of local models, potentially creating a \textit{vicious cycle} of representation inconsistency, classifier divergence, and skewed prototype alignment.
Specifically, representation inconsistency causes classifier updates to diverge and prototype alignment to skew.
Subsequently, these diverged classifiers and skewed prototypes force the feature extractors to map to more inconsistent representation space, thereby exacerbating the vicious cycle.

Motivated by these insights, we propose a novel framework named \textbf{Fed}erated Learning via \textbf{S}emantic \textbf{A}nchors (FedSA) to break the vicious cycle.
Specifically, we first introduce simple yet effective semantic anchors, serving as prototypes, to decouple the generation of prototypes from local representation learning and guide local models in learning consistent representations.
Instead of collecting prototypes from biased models, we project pre-defined class anchors for all categories into the semantic space through a lightweight embedding layer, thereby obtaining semantic anchors that are independent of representation learning and well-separated.
By incorporating semantic anchors, we further propose:
1) Anchor-based Regularization with Margin-enhanced Contrastive Learning (RMCL) to correct the biased feature extractors, enabling them to learn consistent prototypes that exhibit both intra-class compactness and inter-class separability;
2) Anchor-based Classifier Calibration (CC) to correct the biased classifiers, assisting them in learning consistent decision boundaries across different classes.
We then update the semantic anchors with these enhanced prototypes via an Exponential Moving Average (EMA) update, which iteratively encourages clients to collaboratively learn a unified data representation in the semantic space.
Our contributions are summarized as follows:
\begin{itemize}
\item We demonstrate that in prototype-based FL methods, heterogeneity creates a vicious cycle of representation inconsistency, classifier divergence, and skewed prototype alignment across client models.
\item To break the vicious cycle, we propose a novel framework named FedSA that corrects biased feature extractors and classifiers, guiding local models in learning a unified data representation.
\item We evaluate the proposed method under both statistical and model heterogeneity settings. Extensive experiments and ablation studies demonstrate the superiority of FedSA over prototype-based FL methods.
\end{itemize}

\section{Related Work}
\subsection{Model-based FL Approaches}
Model-based FL approaches are inspired by the training scheme that decouples the entire model into a general body and a personalized head \cite{kang2019decoupling, devlin2019bert}.
For example, methods like FedRep \cite{collins2021exploiting}, FedBABU \cite{oh2021fedbabu}, FedRoD \cite{chen2022bridging} and FedGC \cite{niu2022federated} split the model into a feature extractor and a classifier.
This approach allows clients to share a global feature extractor to mitigate data heterogeneity while maintaining client-specific classifiers for personalization.
LG-FedAvg \cite{liang2020think} and FedGen \cite{zhu2021data} treat the top layers as the general body for sharing while allowing the bottom layers to have different architectures.
However, these methods commonly involve significant communication costs as model parameters are transmitted between clients.
Recent studies initialize class anchors \cite{zhou2023fedfa} or use prototypes \cite{xu2023personalized, dai2023tackling}, serving as additional knowledge to align feature representations from different feature extractors.
Nevertheless, these class anchors or prototypes are also updated or generated based on locally inconsistent representations.
In this paper, we only share lightweight semantic anchors across clients in a model-agnostic manner.

\begin{figure*}[ht]
    \!\!\centerline{\includegraphics[width=0.82\linewidth]{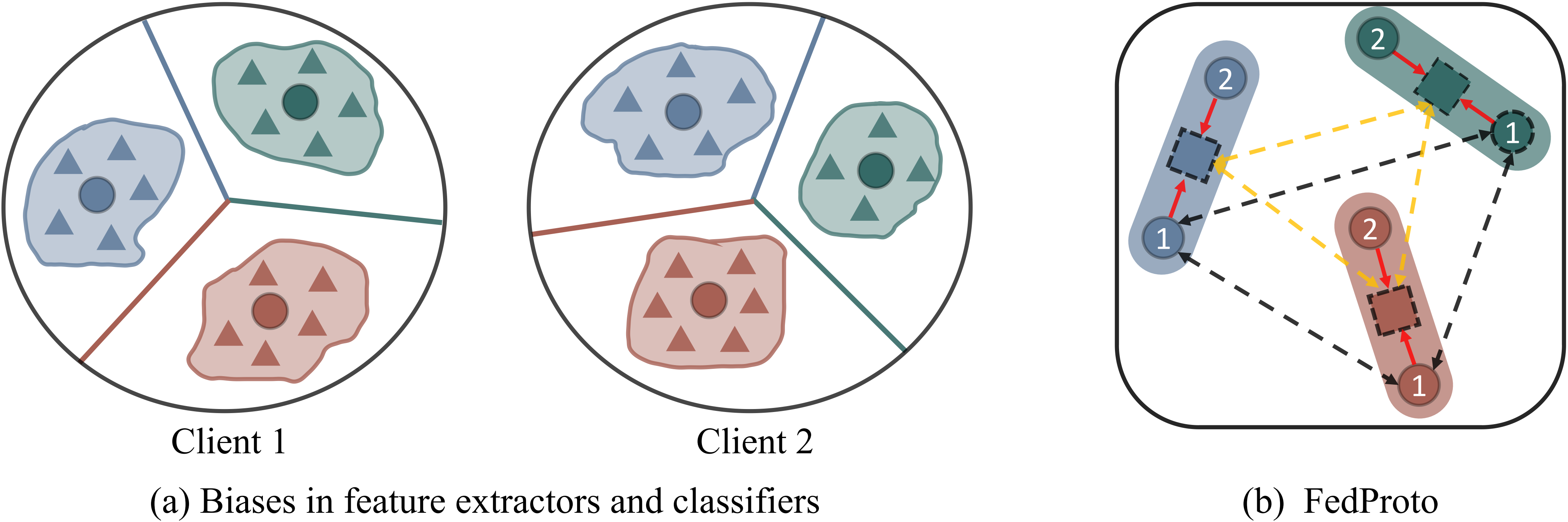}}
    \caption{
    A toy example with two clients shows the vicious cycle.
    Different colors represent classes, while triangles, circles and squares represent data representations, local prototypes and global prototypes, respectively.
    Different colored solid lines indicate decision boundaries.
    Black and yellow dotted arrows show the inter-class separation among prototypes, and red arrows depict the guidance of prototypes during local training.
    Fig.~\ref{fig: motivation}(a) shows that biased datasets lead feature extractors and classifiers to learn inconsistent representations and skewed decision boundaries.
    Fig.~\ref{fig: motivation}(b) shows that in FedProto, the naive averaging aggregation and local guidance reduce the separability of global prototypes and inevitably create a vicious cycle.
    }
    \label{fig: motivation}
\end{figure*}

\subsection{Knowledge-based FL Approaches}
Knowledge-based FL approaches transfer various types of global knowledge instead of model parameters, making them well-suited for model heterogeneity setting.
The typical method in these approaches \cite{jeong2018communication,li2019fedmd,lin2020ensemble} involves incorporating KD methods \cite{hinton2015distilling} to transfer knowledge from a teacher model on the server to enhance the performance of student models on the client side.
FML \cite{shen2020federated} and FedKD \cite{wu2022communication} advance KD methods in a data-free manner, where they share a small auxiliary model as global knowledge, rather than relying on a global dataset.
However, the majority of these methods are not suitable for heterogeneous FL, as their effectiveness heavily relies on the quality of the public dataset or the auxiliary model.
Another popular approach is to transfer abstract class representations, i.e., prototypes, between the server and clients.
FedProto \cite{tan2022fedproto} regularizes the local prototypes to align closer to the aggregated global prototypes.
FedPCL \cite{tan2022federated} employs prototype-wise contrastive learning to enhance the inter-class separability among prototypes.
Additionally, FedTGP \cite{zhang2024fedtgp} treats prototypes as trainable elements and enforces an adaptive margin between them to improve separability.
However, these methods often overlook the biases arising from heterogeneity, which can skew local models toward learning inconsistent representations and potentially create a vicious cycle.
In this paper, we introduce semantic anchors that are independent of local representation learning to guide local models in learning consistent representations.

\section{Method}
\subsection{Problem Statement}
In prototype-based FL settings, we consider a central server and $m$ clients collaboratively training their models with heterogeneous architectures on their heterogeneous local datasets $\left\{\mathcal{D}_i\right\}_{i=1}^m$.
Following FedProto \cite{tan2022fedproto}, we split each client $i$'s model $w_i$ into a feature extractor $f_i$ parameterized by $\theta_i$ and a classifier $h_i$ parameterized by 
$\phi_i \in \mathbb{R}^{C \times D}$, where $D$ is the dimension of the last representations and $C$ is the number of classes.
The goal is for clients to collaborate by sharing global prototypes $\mathcal{P}$ with the central server. The overall collaborative training objective is:
\begin{equation}
\label{eq:1}
\min _{\left\{\left\{\theta_i, \phi_i\right\}\right\}_{i=1}^m} \frac{1}{m} \sum_{i=1}^m \mathcal{L}_i\left(\mathcal{D}_i, \theta_i, \phi_i, \mathcal{P}\right)
\end{equation}

During local training, each client $i$ first computes its local prototype for each class $c$ as follows:
\begin{equation}
\label{eq:2}
P_i^{c}=\frac{1}{\left|\mathcal{D}_{i, c}\right|} \sum_{(x, y) \in \mathcal{D}_{i, c}} f_i\left(x ; \theta_i \right)
\end{equation}
where $\mathcal{D}_{i, c}$ denotes the subset of the local dataset $\mathcal{D}_i$ consisting of all data samples belonging to class $c$. 
After receiving all local prototypes from clients, the server performs weighted-averaging aggregation for each class prototype to obtain the global prototypes:
\begin{equation}
\label{eq:3}
\bar{P}^c=\frac{1}{\left|\mathcal{N}_c\right|} \sum_{i \in \mathcal{N}_c} \frac{\left|\mathcal{D}_{i, c}\right|}{N_c} P_i^c
\end{equation}
where $\mathcal{N}_c$ denotes the set of clients that have class $c$, and $N_c$ represents the total number of data samples of class $c$ among all clients. 
The server then sends the global prototypes $\mathcal{P} = \{\bar{P}^c\}_{c=1}^C$ to each client to guide local training:
\begin{equation}
\label{eq:4}
\mathcal{L}_i=\mathcal{L}_S\left(h_i\left(f_i\left(x ; \theta_i\right) ; \phi_i\right), y\right)+\lambda \mathcal{L}_R\left(P_i^c, \bar{P}^c\right)
\end{equation}
where $\mathcal{L}_S$ is a typical loss term (e.g., cross-entropy loss) in supervised learning.
$\lambda$ is a hyperparameter, and $\mathcal{L}_R$ is a regularization term that minimizes the Euclidean distances between the local prototype $P_i^c$ and the global prototype $\bar{P}^c$.

\begin{figure*}[ht]
\centerline{\includegraphics[width=0.95\linewidth]{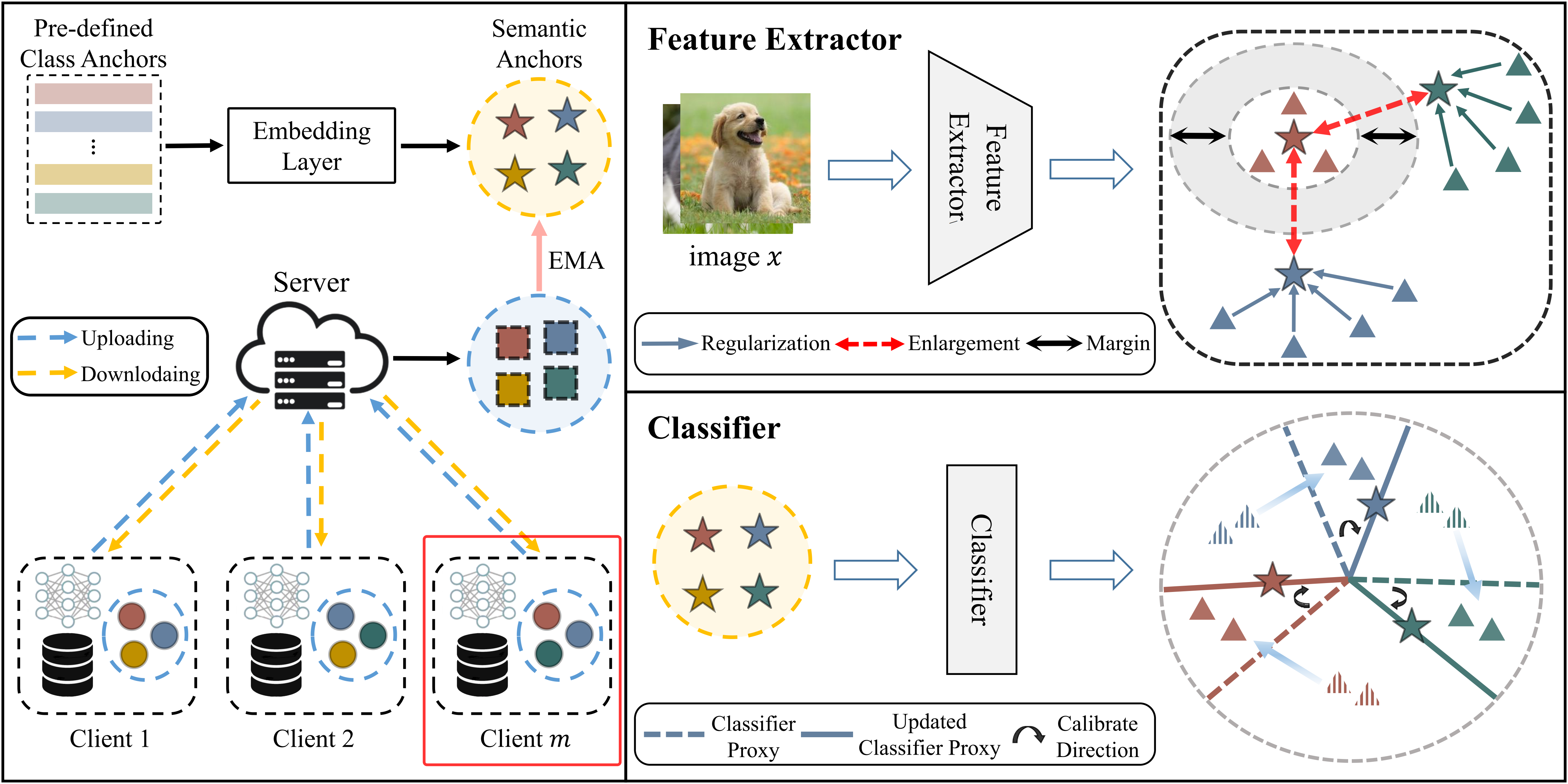}}
    \caption{
    Framework of the proposed method FedSA.
    Pentagons represent semantic anchors.
    In the local training of client $m$, the semantic anchors correct the biased feature extractor and classifier via RMCL and CC, achieving intra-class compactness and inter-class separability of local prototypes, while ensuring consistent decision boundaries.
    On the server, we update the semantic anchors with global prototypes via an EMA update to facilitate collaboration in FL.
    }
    \label{fig: framework}
\end{figure*}

\subsection{Motivation}
Although prototype-based FL methods have achieved significant results by sharing lightweight prototypes, they have overlooked the biases arising from heterogeneity.
Building on previous findings \cite{oh2021fedbabu,guo2023fedbr}, which link the feature extractor to representation learning and the classifier to linear decision boundary learning, we seek to understand how heterogeneity affects the generation of prototypes through these components.
Due to statistical heterogeneity, the dataset of each client is inherently biased.
Consider different local models $w_i$ and $w_j$, each trained on its respective local datasets $\mathcal{D}_i$ and $\mathcal{D}_j$.
As shown in Fig.~\ref{fig: motivation}, we use a toy example to show the existence of biases in local models.

\begin{example}[Representation inconsistency]
Given the same inputs $x$, the outputs $f_i(x)$ could deviate significantly from $f_j(x)$.
Fig.~\ref{fig: motivation}(a) shows the data representations learned by different feature extractors.
As local models are trained on biased datasets, we observe that the representations extracted by different local feature extractors are inconsistent in the semantic space.
\end{example}

\begin{example} [Classifier divergence]
As illustrated in Fig.~\ref{fig: motivation}(a), we visualize the decision boundaries learned by different classifiers $h_i(\cdot)$ and $h_j(\cdot)$ for the same input.
The observation confirms that inconsistent representations significantly cause local classifiers to diverge, resulting in skewed decision boundaries among clients.
\end{example}

\begin{example} [Skewed prototype alignment]
As shown in Fig.~\ref{fig: motivation}(b), although local models can learn well-separated prototypes, representation inconsistency naturally causes these prototypes to align skewed in the semantic space.
The variability in margins between different local prototypes, due to model heterogeneity, further exacerbates this skewed alignment.
Following the naive averaging aggregation, the margins among global prototypes tend to diminish, thus reducing the overall separability and effectiveness of the global prototypes.
Moreover, the global prototypes serve to guide the training of local models.
As indicated by Eq.\ref{eq:4}, the regularization term iteratively adjusts the well-separated local prototypes closer to the less effective global prototypes, impairing the training of client models and forcing them to produce more inconsistent representations.
Therefore, statistical and model heterogeneity potentially creates a vicious cycle of representation inconsistency, classifier divergence, and skewed prototype alignment.

\end{example}

To break this vicious cycle, we introduce simple yet effective semantic anchors to decouple the generation of prototypes from local representation learning and guide local models in learning consistent representations.
Our method is motivated by the fact that prototypes are not only bound to the data distribution but also guided by the objective function in the training paradigm of empirical risk minimization.
By explicitly guiding class representations toward corresponding semantic anchors that are independent of representation learning and well-separated, we can achieve more consistent and discriminative local prototypes.

\subsection{Semantic Anchors}
With a total of $C$ classes in the whole dataset, we randomly initialize pre-defined class anchors $A=\left\{A^c\right\}_{c=1}^C$, where each $A^c \in \mathbb{R}^D$. 
Then, we project them into the semantic space through a trainable embedding layer $h_\psi(A)$ to improve their separability, obtaining semantic anchors $\bar{A}=\left\{\bar{A}^c\right\}_{c=1}^C \in \mathbb{R}^{C \times D}$.
In each communication round $t$, the server sends these semantic anchors $\bar{A}^t$ to each client to guide their local training. 
The goal is for clients to collaboratively learn a unified data representation that exhibits intra-class compactness and inter-class separability via the FL paradigm.
To achieve this, we further propose Anchor-based Regularization with Margin-enhanced Contrastive Learning (RMCL) and Anchor-based Classifier Calibration (CC).
An overview of the proposed framework is shown in Fig.~\ref{fig: framework}.

\noindent\textbf{Anchor-based regularization.}
To achieve intra-class compactness, we propose anchor-based regularization by directly minimizing the distance between class representations and corresponding semantic anchors.
In Eq.\ref{eq:4}, we redefine the regularization loss as follows:
\begin{equation}
\mathcal{L}_R=\sum_{c \in\mathcal{D}_i} d\left(P_i^c, \bar{A}^c\right)
\end{equation}
where $d$ measures the Euclidean distance between the local prototype $P_i^c$ and the semantic anchor $\bar{A}^c$.
Although guiding class representations toward corresponding semantic anchors can achieve an intra-class compact embedding space, due to statistical and model heterogeneity, clients often extract highly diverse data representations that vary in separability and prototype margins across different classes.

\noindent\textbf{Anchor-based margin-enhanced contrastive learning.}
Inspired by previous works \cite{deng2019arcface,chen2021large,zhang2024fedtgp}, we propose anchor-based margin-enhanced contrastive learning that enforces a client-specific margin $d_i^*$ between classes to facilitate the learning of a large-margin data representation and ensure consistent prototype margins across clients.
Specifically, we calculate the average margin among semantic anchors, referred to as the global margin $d_{global}^t$, and the average margin among local prototypes for each client, referred to as the local margin $d_i^t$. 
For client $i$, the local margin is computed:
\begin{equation}
d_i^t=\frac{1}{(N-1)^2} \sum_a \sum_{b \neq a} d\left(P_i^a, P_i^b\right)
\end{equation}
where $N$ is the number of classes maintained by client $i$.
The client-specific margin $d_i^*$ for each client is determined by taking the larger value between the local and global margins, i.e., $d_i^*=\max \left\{d_{global}^t, d_i^t\right\}$.
Our goal is to decrease the distance between $P_i^c$ and $\bar{A}^c$, and to increase the distance between $P_i^c$ and $\bar{A}^{c^{\prime}}$ with the client-specific margin $d_i^*$.
Accordingly, we define the anchor-based margin-enhanced contrastive loss as follows: 
\begin{equation}
\mathcal{L}_{MCL}=-\log \frac{e^{-\left(d\left(P_i^c, \bar{A}^c\right)+d_i^*\right)}}{e^{-\left(d\left(P_i^c, \bar{A}^c\right)+d_i^*\right)}+\sum_{c^{\prime}} e^{-d\left(P_i^c, \bar{A}^{c^{\prime}}\right)}}
\end{equation}
where $c^{\prime} \in[C]$, $c^{\prime} \neq c$.
The client-specific margin adaptively guides each client toward better data representations and consistent prototype margins.
Specifically, when the local model learns well-separated and clustered class representations, the client adopts its local margin $d_i^t$ as the client-specific margin.
Otherwise, the client uses the global margin $d_{global}^t$ to encourage the local model to learn class representations with a larger margin separation.
Through repeated collaborative training in FL, clients reach a consensus on the margins among local prototypes.
By aggregating these uniformly adjusted local prototypes at the server, we can obtain global prototypes with enhanced separability.

\noindent\textbf{Anchor-based classifier calibration.}
Beyond correcting biased feature extractors to learn consistent representations, we also propose anchor-based classifier calibration to correct biased classifiers toward achieving consistent decision boundaries.
Specifically, each client $i$ uses the semantic anchors as inputs for its classifier, calibrating the classifiers according to the following objective:
\begin{equation}
\mathcal{L}_{CC}=-\frac{1}{C} \sum_{c \in C} \log \frac{e^{\left( \phi_{i, c}^{\top} \bar{A}^c \right)}}{\sum_{j=1}^C e^{\left( \phi_{i, j}^{\top} \bar{A}^c \right)}}
\end{equation}
where $\mathcal{L}_{CC}$ is the classifier calibration loss, which reduces the distance between the $c$-th class proxy and the corresponding semantic anchor.
By applying this loss across different clients, we not only reduce the discrepancies in decision boundaries between different model architectures but also promote the learning of consistent representations.

\noindent\textbf{Overall.}
By integrating all components, the semantic anchors effectively guide each client to learn a unified data representation by minimizing the overall loss:
\begin{equation}
\label{eq:9}
\mathcal{L}_i=\mathcal{L}_S+\lambda_1 \mathcal{L}_R+\lambda_2 \mathcal{L}_{MCL}+\lambda_3 \mathcal{L}_{CC}
\end{equation}

\begin{table*}[ht]
  \centering
  \renewcommand{\arraystretch}{0.9}
  \resizebox{\linewidth}{!}{
    \tiny
    \begin{tabular}{l|*{3}{c}|*{3}{c}}
    \toprule
    Settings & \multicolumn{3}{c|}{Cross-silo Setting} & \multicolumn{3}{c}{Cross-device Setting} \\
    \midrule
    Datasets & Cifar10 & Cifar100 & Tiny-ImageNet & Cifar10 & Cifar100 & Tiny-ImageNet \\
    \midrule
    LG-FedAvg & 89.32$\pm$0.04 & 48.97$\pm$0.07 & 35.08$\pm$0.06 & 87.27$\pm$0.05 & 43.41$\pm$0.10 & 28.54$\pm$0.05 \\
    FedGen & 89.13$\pm$0.05 & 48.25$\pm$0.06 & 32.68$\pm$0.08 & 86.99$\pm$0.07 & 42.70$\pm$0.08 & 27.76$\pm$0.08 \\
    FML & 89.42$\pm$0.03 & 48.73$\pm$0.22 & 34.75$\pm$0.13 & 87.42$\pm$0.07 & 42.13$\pm$0.14 & 27.63$\pm$0.04 \\
    FedKD & 90.21$\pm$0.06 & 53.14$\pm$0.10 & 37.14$\pm$0.16 & 87.94$\pm$0.13 & 46.03$\pm$0.15 & 30.58$\pm$0.19 \\
    FedDistill & 89.61$\pm$0.03 & 49.27$\pm$0.18 & 35.45$\pm$0.07 & 87.57$\pm$0.05 & 42.15$\pm$0.16 & 28.57$\pm$0.12 \\
    FedProto & 90.76$\pm$0.05 & 53.26$\pm$0.11 & 36.23$\pm$0.12 & 87.49$\pm$0.11 & 44.46$\pm$0.32 & 28.73$\pm$0.21 \\
    FedTGP & 90.23$\pm$0.08 & 53.71$\pm$0.13 & 36.43$\pm$0.08 & 87.98$\pm$0.16 & 47.41$\pm$0.11 & 29.96$\pm$0.17 \\
    \midrule
    FedSA & \textbf{90.88$\pm$0.04} & \textbf{54.39$\pm$0.12} & \textbf{37.30$\pm$0.13} & \textbf{88.78$\pm$0.08} & \textbf{48.42$\pm$0.17} & \textbf{30.82$\pm$0.14} \\
    \bottomrule
    \end{tabular}}
    \caption{
  The test accuracy (\%) on three datasets in both cross-silo and cross-device settings under statistical heterogeneity.
  }
    \label{tab:statistical heterogeneity acc}
\end{table*}

\subsection{FedSA Framework}
Following the prototype-based communication protocol, each client collects local prototypes and sends them to the server. 
The server then performs weighted-averaging aggregation to obtain global prototypes.
To facilitate collaboration in FL, we update the semantic anchors with these global prototypes using the EMA update as follows:
\begin{equation}
\label{eq:10}
\bar{A}^{t+1}=\alpha \bar{A}^{t}+(1-\alpha) \bar{P}^{t}
\end{equation}
where $\alpha$ is a decay factor that controls the update rate.
Similar to prototype-based FL methods, FedSA transmits only compact 1D-class semantic anchors, which naturally bring benefits for both privacy preservation and communication efficiency.
Moreover, by integrating the anchor-based components and the EMA update, FedSA builds a positive feedback loop that iteratively encourages clients to collaboratively learn a unified data representation that exhibits intra-class compactness and inter-class separability.

\section{Experiments}
\subsection{Setup}
\noindent\textbf{Datasets.} 
We consider image classification tasks and evaluate our method on three popular datasets, including Cifar10, Cifar100 \cite{deng2009imagenet}, and Tiny-Imagenet \cite{chrabaszcz2017downsampled}.

\noindent\textbf{Baseline methods.}
To evaluate our proposed method FedSA, we consider both statistical and model heterogeneous settings and compare it with popular methods including LG-FedAvg \cite{liang2020think}, FedGen \cite{zhu2021data}, FML \cite{shen2020federated}, FedKD \cite{wu2022communication}, FedDistill \cite{jeong2018communication}, FedProto \cite{tan2022fedproto}, and FedTGP \cite{zhang2024fedtgp}.

\noindent\textbf{Statistical heterogeneity.}
Like previous studies \cite{li2021model, li2022federated}, we simulate statistical heterogeneity among clients by using Dirichlet distribution.
Specifically, we first sample $q_{c, i} \sim \operatorname{Dir}(\beta)$ for each class $c$ and client $i$.
We then allocate the proportion $q_{c, i}$ of data points from class $c$ in the dataset to client $i$, where $\operatorname{Dir}(\beta)$ is the Dirichlet distribution with a concentration parameter $\beta$ set to 0.1 by default.
All clients use a 4-layer CNN \cite{mcmahan2017communication} for a homogeneous model setting.

\noindent\textbf{Model heterogeneity.}
Following FedTGP \cite{zhang2024fedtgp}, we simulate model heterogeneity among clients by assigning different model architectures.
Specifically, we denote this setting as "HtFE$_X$", where FE$_X$ represents the $X$ distinct feature extractors used. 
Each client $i$ is assigned the ($i$ mod $X$)th model architecture.
For instance, the "HtFE$_8$" setting includes eight architectures: 4-layer CNN, GoogleNet \cite{szegedy2015going}, MobileNet\_v2 \cite{sandler2018mobilenetv2}, ResNet18, ResNet34, ResNet50, ResNet101, and ResNet152 \cite{he2016deep}, used in our main experiments.
To generate feature representations with an identical feature dimension $K$, an average pooling layer is added after each feature extractor, with $K$ set to 512 by default. 

\noindent\textbf{Implementation details.}
We consider two widely used FL settings, the cross-silo setting and the cross-device setting \cite{kairouz2021advances}.
In the cross-silo setting, we set the total number of clients to 20 with a client participation ratio $\rho$ of 1.
In the cross-device setting, the total number of clients is set to 100 with a participation ratio $\rho$ of 0.1.
Our main experiments are conducted in the cross-device setting, where the participating clients are generally resource-constrained edge devices, more closely reflecting real-world FL scenarios.

Unless explicitly specified, we follow previous studies \cite{zhang2023pfllib} by running one local epoch of training on each client per round, using a batch size of 10 and a learning rate $\lambda$ = 0.01 for 1000 communication rounds.
Each client’s local dataset is split into a training set (75\%) and a test set (25\%), with the test set used to evaluate performance.
We run three trials for all experiments and report the mean and standard deviation.
For our FedSA, we set $\lambda_1 = 0.1$ (the same as in prototype-based FL methods), $\lambda_3 = 1$, and $\alpha = 0.9999$.
We use smaller $\lambda_2 = 0.01$ for statistical heterogeneity, and $\lambda_2 = 1$ for model heterogeneity.
Please refer to the Appendix for more results and details.

\subsection{Performance}
\noindent\textbf{Impact of statistical heterogeneity.}
Table~\ref{tab:statistical heterogeneity acc} shows the accuracy of all methods in both cross-silo and cross-device settings under statistical heterogeneity.
Comparing different FL methods, we observe that FedSA consistently outperforms all other methods across all tasks.
Specifically, FedSA outperforms the baselines on three datasets by up to 6.29\%.
In more complex tasks, the increased number of classes leads to more skewed prototypes during aggregation, exacerbating the vicious cycle.
FedSA improves prototype-based FL methods by up to 3.96\% through the incorporation of lightweight but innovative semantic anchors serving as prototypes.
This improvement demonstrates the effectiveness of semantic anchors in addressing statistical heterogeneity, enabling local clients to learn consistent representations.

\noindent\textbf{Impact of model heterogeneity.}
Table~\ref{tab:model heterogeneity acc} shows the accuracy on Cifar100 in the cross-device setting under both statistical and model heterogeneity. 
As the number of feature extractors $X$ increases, we observe a significant degradation in the performance of prototype-based FL methods compared to other methods.
This confirms that prototype-based methods heavily rely on the quality of data representation, and the naive aggregation of prototypes with varying margins, such as FedProto, leads to catastrophic performance degradation.
Although FedTGP enhances the separability of prototypes through margin-enhanced contrastive learning, it fails to achieve better performance due to its neglect of biases in local clients.
Using our RMCL and CC, FedSA can improve FedProto by up to 19.37\% and FedTGP by up to 3.78\%, demonstrating that FedSA is more robust and less affected by model heterogeneity.

\begin{table}[t]
  \centering
  \renewcommand{\arraystretch}{1.1}
  \huge
  \resizebox{\linewidth}{!}{
    \begin{tabular}{l|*{4}{c}}
    \toprule[2pt]
    Settings & HtFE$_2$ & HtFE$_3$ & HtFE$_5$ & HtFE$_8$ \\
    \midrule
    LG-FedAvg & 41.22$\pm$0.13 & 40.59$\pm$0.42 & 39.66$\pm$0.23 & 35.24$\pm$0.13 \\
    FedGen & 40.05$\pm$0.18 & 38.47$\pm$0.29 & 39.48$\pm$0.28 & 34.63$\pm$0.44 \\
    FML & 40.62$\pm$0.06 & 40.19$\pm$0.09 & 36.39$\pm$0.13 & 34.47$\pm$0.09 \\
    FedKD & 42.19$\pm$0.09 & 40.53$\pm$0.07 & 37.47$\pm$0.21 & 35.08$\pm$0.12 \\
    FedDistill & 41.58$\pm$0.15 & 40.92$\pm$0.36 & 39.98$\pm$0.13 & 35.26$\pm$0.48 \\
    FedProto & 35.63$\pm$0.22 & 29.89$\pm$0.48 & 27.75$\pm$0.64 & 16.31$\pm$0.32 \\
    FedTGP & 42.14$\pm$0.29 & 37.97$\pm$0.57 & 38.66$\pm$0.42 & 32.04$\pm$0.74 \\
    \midrule
    FedSA & \textbf{43.54$\pm$0.21} & \textbf{41.75$\pm$0.16} & \textbf{40.96$\pm$0.12} & \textbf{35.68$\pm$0.17} \\
    \bottomrule[2pt]
    \end{tabular}}
    \caption{
  The test accuracy (\%) on Cifar100 in the cross-device setting under both statistical and model heterogeneity.
  }
    \label{tab:model heterogeneity acc}
\end{table}

\subsection{Number of Local Epochs}

\begin{table}[t]
  \centering
  \renewcommand{\arraystretch}{0.9}
  \resizebox{\linewidth}{!}{
    \begin{tabular}{l|*{3}{c}}
    \toprule
     & $E=1$ & $E=5$ & $E=10$ \\
    \midrule
    LG-FedAvg   & 35.24$\pm$0.13 & 34.98$\pm$0.15 & 34.78$\pm$0.09 \\
    FedGen      & 34.63$\pm$0.44 & 33.89$\pm$0.39 & 32.95$\pm$0.06 \\
    FML         & 34.47$\pm$0.09 & 35.22$\pm$0.05 & 35.02$\pm$0.31 \\
    FedKD       & 35.08$\pm$0.12 & 37.45$\pm$0.57 & 37.16$\pm$0.11 \\
    FedDistill  & 35.24$\pm$0.48 & 36.15$\pm$0.21 & 36.17$\pm$0.17 \\
    FedProto    & 16.31$\pm$0.32 & 23.96$\pm$0.55 & 22.88$\pm$0.45 \\
    FedTGP      & 32.04$\pm$0.74 & 34.86$\pm$0.82 & 34.51$\pm$0.25 \\
    \midrule
    FedSA & \textbf{35.68$\pm$0.17} & \textbf{37.57$\pm$0.27} & \textbf{37.43$\pm$0.13} \\
    \bottomrule
    \end{tabular}}
    \caption{The test accuracy (\%) with different number of local epochs on Cifar100 in the HtFE$_8$ setting.}
    \label{tab:num_local_epochs}
\end{table}

Clients can perform more local epochs to alleviate communication costs, albeit at the expense of exacerbating the biases across clients \cite{wang2020tackling}.
The results of different local epochs are shown in Table~\ref{tab:num_local_epochs}.
We set the number of communication rounds to 500 for $E=5$ and 200 for $E=10$.
We note that as the number of local epochs increases, the performance of model-based methods declines, whereas knowledge-based methods show improvement.
This improvement is attributed to the increased number of local epochs enhancing the model’s capability to extract more fine-grained knowledge.
There remains a significant performance gap between prototype-based FL methods and other knowledge-based methods.
In contrast, our method FedSA maintains better performance, demonstrating its effectiveness with different numbers of local epochs.

\subsection{Communication Efficiency}


\begin{figure}[t]
	\centering
	\includegraphics[width=0.84\linewidth]{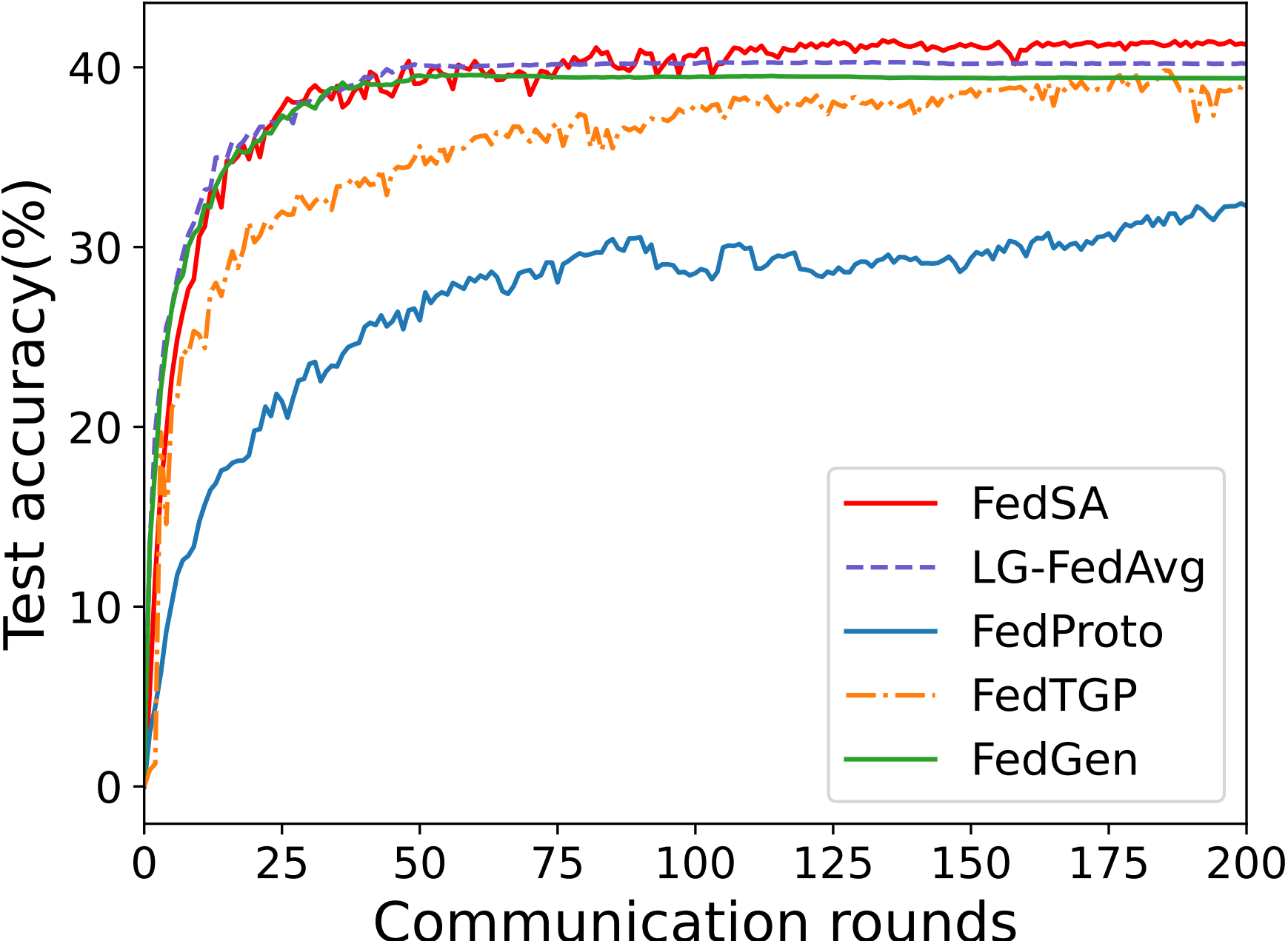}
	\caption{The test accuracy (\%) on Cifar100 in the cross-silo setting using the HtFE$_8$ setting under both statistical and model heterogeneity.} 
        \label{fig:com}
\end{figure}

Fig.~\ref{fig:com} shows the accuracy in the cross-silo setting using the HtFE$_8$ settings under both statistical and model heterogeneity.
We observe that the model-based method LG-FedAvg and the KD-based method FedGen show faster convergence compared to prototype-based methods. 
The convergence curve of prototype-based methods FedProto and FedTGP initially exhibits significant volatility, due to the inadequate feature extraction capabilities of heterogeneous models in the early stages of training.
Moreover, prototypes are more susceptible to the impacts of statistical and model heterogeneity, resulting in lower accuracy for prototype-based methods.
By guiding class representations toward well-separated semantic anchors, FedSA can effectively increase accuracy without slowing down convergence, while also ensuring privacy preservation.

\subsection{Ablation and Hyperparameter Studies}
In Table~\ref{tab:ab}, we evaluate the efficacy of each component in our method on Cifar100 in the HtFE$_8$ setting.
In FedSA, EMA and $\mathcal{L}_{R}$ serve as default components.
Without embedding projection, data representations are directly regularized by randomly initialized, pre-defined class anchors.
The accuracy surpasses that of FedProto by 13.23\%, validating our motivation to decouple the generation of prototypes from local representation learning.
To enhance the separability of semantic anchors, we project pre-defined class anchors through an embedding layer and guide feature extractors to learn consistent and large margins between prototypes using $\mathcal{L}_{MCL}$. 
This approach can improve performance by up to 3.28\% and outperforms FedTGP.
Furthermore, we correct biased classifiers by utilizing $\mathcal{L}_{CC}$ to ensure consistent decision boundaries.
Combining all components, FedSA guides local clients to learn a unified data representation, achieving a significant improvement of 6.14\% over the baseline.
We also notice that FedSA without $\mathcal{L}_{MCL}$ shows a marginal improvement in performance, highlighting the importance of learning consistent representations.

Table~\ref{tab:hyper} shows the accuracy of our method with varying hyperparameters $\lambda_2$ and $\lambda_3$.
FedSA performs better with $\lambda_2$ ranging from 1 to 5.
However, there is a slight drop in accuracy when $\lambda_2 = 10$.
This drop is attributed to a larger $\lambda_2$ continuously expanding the margins among prototypes, where overly large margins result in unstable prototype guidance for clients.
For $\lambda_3$, the highest accuracy is achieved at 1, but increasing this hyperparameter leads to excessive classifier optimization, which compromises the model's generalizability and significantly degrades performance.
Notably, even with $\lambda_2 = 0.1$ or $\lambda_3 = 0.1$, FedSA still achieves an accuracy of at least 33.26\%, outperforming other prototype-based FL methods.

\begin{table}[t]
    \centering
    \tiny
    \renewcommand{\arraystretch}{1}
    \resizebox{1\linewidth}{!}
    {
    \begin{tabular}{ccccc|c}
    \toprule[1pt]
     EMA  & $\mathcal{L}_{R}$  & ER & $\mathcal{L}_{MCL}$ & $\mathcal{L}_{CC}$ & Acc (\%) \\ 		
     \midrule[0.5pt]
     \midrule[0.5pt]
    $\checkmark$ & $\checkmark$ & & & & 29.54\\ 
    \midrule
    $\checkmark$ & $\checkmark$ & $\checkmark$ &  &   & 30.80 ({\textbf{+1.26}})\\ 
    $\checkmark$ & $\checkmark$ & $\checkmark$ &  & $\checkmark$ & 30.61 ({\textbf{+1.07}})\\ 
    $\checkmark$ & $\checkmark$ & $\checkmark$ & $\checkmark$ & & 32.82 ({\textbf{+3.28}})\\ 
    $\checkmark$ & $\checkmark$ & $\checkmark$ & $\checkmark$ & $\checkmark$& 35.68 ({\textbf{+6.14}})\\ 
    \bottomrule[1pt]
    \end{tabular}
     }
    \caption{Ablation studies on the key components of our proposed FedSA on Cifar100 in the HtFE$_8$ setting. $\mathcal{L}_{R}$: Anchor-based Regularization. $ER$: Embedding Projection. $\mathcal{L}_{MCL}$: Anchor-based Margin-enhanced Contrastive Learning. $\mathcal{L}_{CC}$: Anchor-based Classifier Calibration.}
    \label{tab:ab}
\end{table}

\begin{table}[t]
  \centering
  \resizebox{\linewidth}{!}{
    \begin{tabular}{l|cccc|cccc}
    \toprule
     & \multicolumn{4}{c|}{$\mathcal{L}_{MCL}$} & \multicolumn{4}{c}{$\mathcal{L}_{CC}$} \\
     \midrule
     & $0.1$ & $1$ & $5$ & $10$ & $0.1$ & $1$ & $5$ & $10$ \\
     \midrule
     Acc (\%) & 33.42 & \textbf{35.68} & 35.54 & 35.12 & 33.26 & \textbf{35.68} & 34.66 & 34.29 \\
    \bottomrule
    \end{tabular}}
     \caption{The test accuracy (\%) on Cifar100 in the  HtFE$_8$ setting with different $\lambda_2$ or $\lambda_3$. We set $\lambda_2=1$ and $\lambda_3=1$ by default.}
     \label{tab:hyper}
\end{table}

\section{Conclusion}
In this paper, we propose a novel framework named FedSA, which introduces simple yet effective semantic anchors to decouple the generation of prototypes.
By incorporating semantic anchors, we further propose RMCL and CC to correct local models, enabling them to learn consistent representations and decision boundaries.
We then update the semantic anchors with global prototypes via an EMA update to encourage clients to learn a unified data representation.
Extensive experiments and ablation studies demonstrate the superiority of FedSA over prototype-based FL methods.

\section{Acknowledgments}
This work was supported by Chongqing Natural Science Foundation Innovation and Development Joint Fund (CSTB2023NSCQ-LZX0109) and Fundamental Research Funds for the Central Universities (No.2023CDJQCZX-001). 
The authors thank Jinghua Zhou and Ping Jiang for their invaluable support.

\bibliography{aaai25}

\end{document}